\pdfoutput=1
\documentclass[11pt]{article}

\usepackage[]{EMNLP2023}

\usepackage{times}
\usepackage{latexsym}

\usepackage{algorithm}
\usepackage{algpseudocode}

\usepackage{hyperref}
\usepackage{amsmath}
\usepackage{titlesec}
\usepackage{xcolor}         % colors
\usepackage{graphicx,amssymb,amsmath,bm}
\usepackage[export]{adjustbox}
\usepackage{multirow}
\usepackage{booktabs}
\usepackage{multirow}
\usepackage{svg}

\usepackage[T1]{fontenc}
\usepackage[utf8]{inputenc}

\usepackage{microtype}

\usepackage{inconsolata}

\newcommand{\sys}{Nece}

\title{\textsc{\sys}: Narrative Event Chain Extraction Toolkit}

\author{Guangxuan Xu$^{1,2}$\thanks{\ \ Equal contribution. } , Paulina Toro Isaza$^1$\footnotemark[1] , Moshi Li$^3$\footnotemark[1]\\ \textbf{Akintoye Oloko$^1$, Bingsheng Yao$^4$, Cassia Sanctos$^1$, Aminat Adebiyi$^1$} \\
\textbf{Yufang Hou$^1$, Nanyun Peng$^2$,   Dakuo Wang$^1$} \\
  $^1$ IBM Research \quad
  $^2$University of California, Los Angeles \\
  $^3$Northeastern University  \quad
  $^4$Rensselaer Polytechnic Institute  \\
  \small \tt \{GX.Xu, ptoroisaza, YHou, Dakuo.Wang\}@ibm.com \\
}
\begin{document}
\maketitle
\begin{abstract}

To understand a narrative, it is essential to comprehend the temporal event flows, especially those associated with main characters; however, this can be challenging with lengthy and unstructured narrative texts. To address this, we introduce \textsc{\sys}, an open-access, document-level toolkit that automatically extracts and aligns narrative events in the temporal order of their occurrence. Through extensive evaluations, we show the high quality of the \textsc{\sys} toolkit and demonstrates its downstream application in analyzing narrative bias regarding gender. We also openly discuss the shortcomings of the current approach, and potential of leveraging generative models in future works. Lastly the \textsc{\sys} toolkit includes both a Python library and a user-friendly web interface, which offer equal access to professionals and layman audience alike, to visualize event chain, obtain narrative flows, or study narrative bias. 

\end{abstract}

\section{Introduction}
\label{sec:intro}

In this paper, we introduce \textsc{Nece}, a document-level narrative event chain extraction system with an easily accessible online interface. 
\textsc{\sys} offers visualization of structured event chains from unstructured text corpus, making it easier for readers to access and comprehend information from narrative sources such as fairy-tales, news reports, and memoirs. While there are existing text processing approaches, such as temporal dependency graph parsing~\citep{mathur-etal-2022-doctime}, document-level AMR (abstract meaning representation) parsing ~\citep{naseem-etal-2022-docamr}, as well as scripts and event schema ~\citep{chambers-jurafsky-2008-unsupervised, Dror2022ZeroShotOE}, none offers a end-to-end visualization of the narrative flow as \textsc{Nece} shown in Figure~\ref{fig:demo_character}. The narrative event chain of \textsc{Nece} is a \textit{salience-filtered, temporal-ordered, linear chain of events organized for different gender and character groups.}  We provide a detailed and document level temporal chain, versus the more abstractive intentions of ~\citet{chambers-jurafsky-2008-unsupervised}.

\begin{figure}[tbp]
    \centering
    \includegraphics[width=0.48\textwidth]{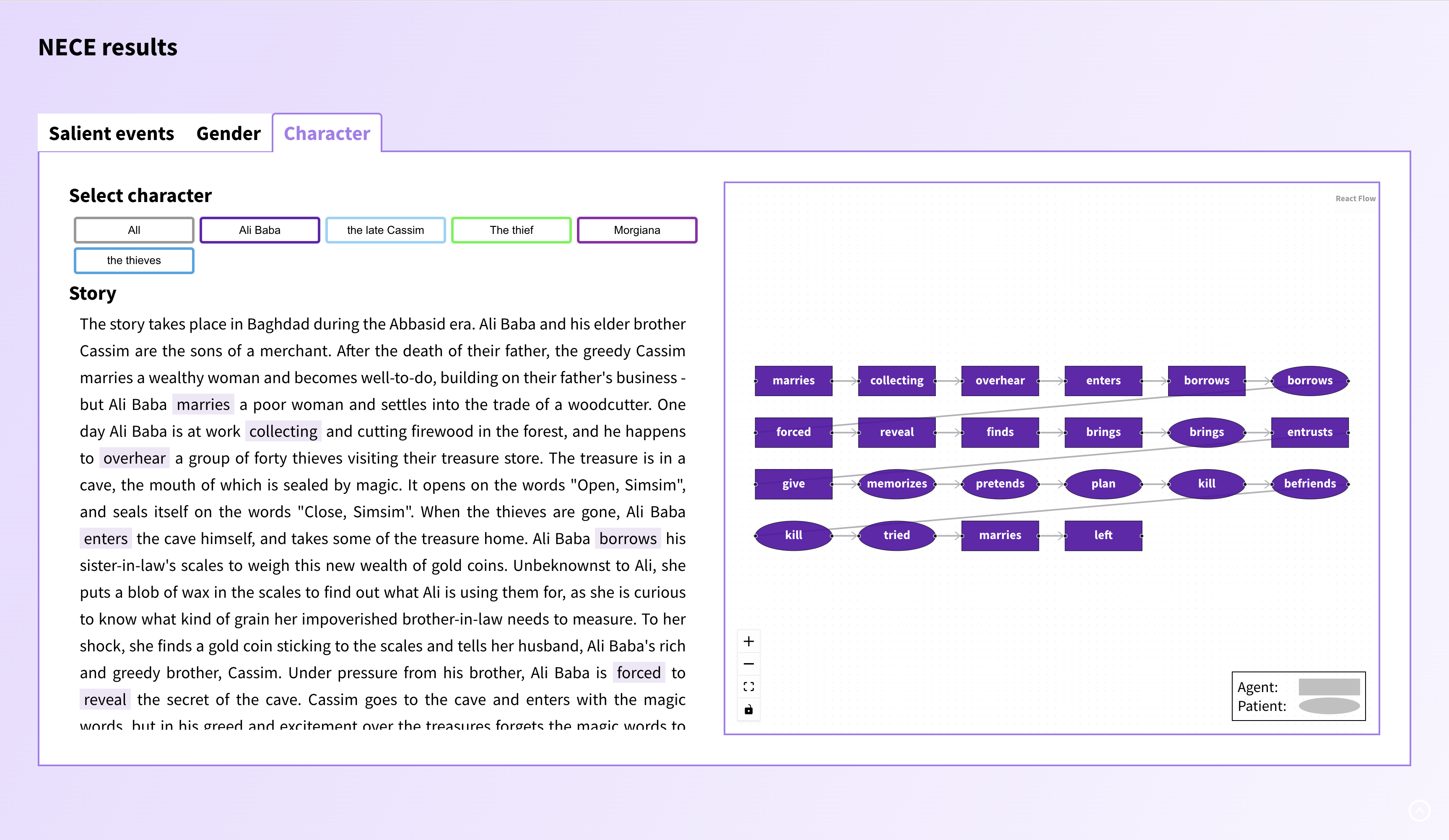}
    \caption{The character interface provides chain of events for main characters of story;}
    \vspace{-0.15in}
    \label{fig:demo_character}
\end{figure}
% This work fills the void by proposing an event chain representation of text, and renders an interface that allows easy access to key temporal and semantic features of any narrative of the user's choice. We choose to use Narrative Event Chain~\citep{chambers-jurafsky-2008-unsupervised,Zhang2021SalienceAwareEC} as the abstract/summarized representation of texts; As defined in \citep{chambers-jurafsky-2008-unsupervised}, narrative event chain has a linear schema made of "partially ordered set of events associated with a protagonist." We expand on the concept to include events associated with all major characters in the narrative, and adopts the salient event concept proposed in ~\citep{Zhang2021SalienceAwareEC} to filter out generic and distracting events. 

% What is narrative Event Chain here? difference from others, especially scripts?

% A narrative event chain representation is advantageous for its narrative-central formulation and clearness for visualization. Its linear chain structure not only conveys the temporal order of event occurrence, which is crucial for narrative understanding, but also highlights the actions(events) and actors(characters) that are the backbone of narrative. Compared with parsing graphs, event chain is much more intuitive and readable; and compared with normal script and event-schema, narrative event chain is more optimized for narrative, with events filtering, character clustering, and temporal ordering capabilities. 

% what's the benefit of narrative event chains?

In order to construct a chain of narrative events, we devised narrative feature extraction algorithms to mine salient events and associate them with characters and gender groups (Section \ref{sec: method}). %gender association, and character association of events.
Salience score is computed as combination of tf-idf score(punishes generic events and rewards locally high-frequency events), character mention score, and location mention score. We also designed a gender identification algorithm by leveraging document-level co-reference resolution results to find gendered names or pronouns.  For the temporal ordering task, we leveraged the SOTA temporal relation (TempRel) model, ECONET~\citep{Han2021ECONETEC}, to predict adjacent pairwise ordering, and devised greedy and window-sliding approaches  to generate ordered chains of events from pairwise relations. 

% We also improved BookNLP's~\cite{Bamman2014ABM} character identification and clustering algorithm to determine character importance by combined name mentions and pronoun mentions.

% Our toolkit is supported by three main capabilities: 1) extracting and filtering events, 2) extracting narrative features, including main characters, gender of characters, and linking characters to events, and 3) ordering the event chain. We built our pipeline leveraging State of the Art existing tools and novel algorithms. For example, BookNLP~\cite{Bamman2014ABM} package, a toolkit built for document-level narratives, is used for correference-resolution and character clustering. We also apply advanced temporal relation classifiers~\citep{Han2021ECONETEC} to order event chains; moreover, we also devised our own gender identification algorithm that supports gender bias detection in event chains. 

% what are main components of the toolkit?

In summary, this work contributes in the following respects: \textbf{First}, we propose a version of narrative event chain that offers a detailed visualization of narrative event flow, suitable for event understanding and analysis. \textbf{Second} We integrate capabilities of excellent toolkits, such as BookNLP~\cite{Bamman2014ABM} and ~\textsc{Econet}, and design novel custom algorithms to build the proposed event chain construction pipeline with validated performance. \textbf{Third}, we open-source the pip package of (\textsc{Nece}) to extract and visualize document-level narrative event chains, and present our website for non-expert users\footnote{The \textsc{Nece} toolkit is licensed under the MIT license and can be accessed \href{https://d3tvz53x8u8tnu.cloudfront.net}{here}; screenshot of demo is \href{https://drive.google.com/file/d/1_H599ClNZ0ee8pD7XM_wqW_nzZMmTZ7b/view?usp=sharing}{here}}. An experimental application in temporal-based bias analysis sheds light on the application potential of \textsc{Nece} toolkit. \textbf{Lastly}, we discuss the technical challenges of existing methods, and explores few-shot temporal reasoning capability of recent popular LLMs.

\section{Related Work}
\label{sec:related-work}

\noindent \textbf{Narrative Event Chain} is inspired by works that model text for different specialty and usage; \citet{Pichotta2015StatisticalSL} uses scripts, linear chains of events, to \textit{model stereotypical sequences of events}~; there are also hierarchical scripts~\citep{Dror2022ZeroShotOE} , which creates separate event chains for each detected hierarchy; besides linear chains, document-level semantic graphs ~\cite{naseem-etal-2022-docamr} can offer more detailed view of text content;

\noindent \textbf{Salience detection} is an important detail that underpins the utility power of our narrative chain. Different from salience algorithm in \citet{jindal-etal-2020-killed, Zhang2021SalienceAwareEC}, which extract summarizing/abstrative events by training on news summarization corpus; Our salience filtering only aims to remove generic words, auxiliary verbs, and non-narrative related events, while the bulk of event content shall be kept. Salience filtering is the pre-processing step that reduces distractions for temporal ordering model, and render the resulting narrative flow more readable and useful.

% Recent work ~proposes a hierarchical script schema to represent narratives, which implies creating a separate temporal event chain for each detected hierarchy. In addition to linear scripts, graphs are also common structures for narrative event representation. For example, DocAMR~ parses an entire document into a large and detailed semantic graph. ~\newcite{Huang2016LiberalEE} extracts an event schema from unstructured text by grouping event triggers into high-level categories for the purpose of detecting patterns and opportunities in text. 

\begin{figure}[tbp]
    \centering
    \includegraphics[width=0.48\textwidth]{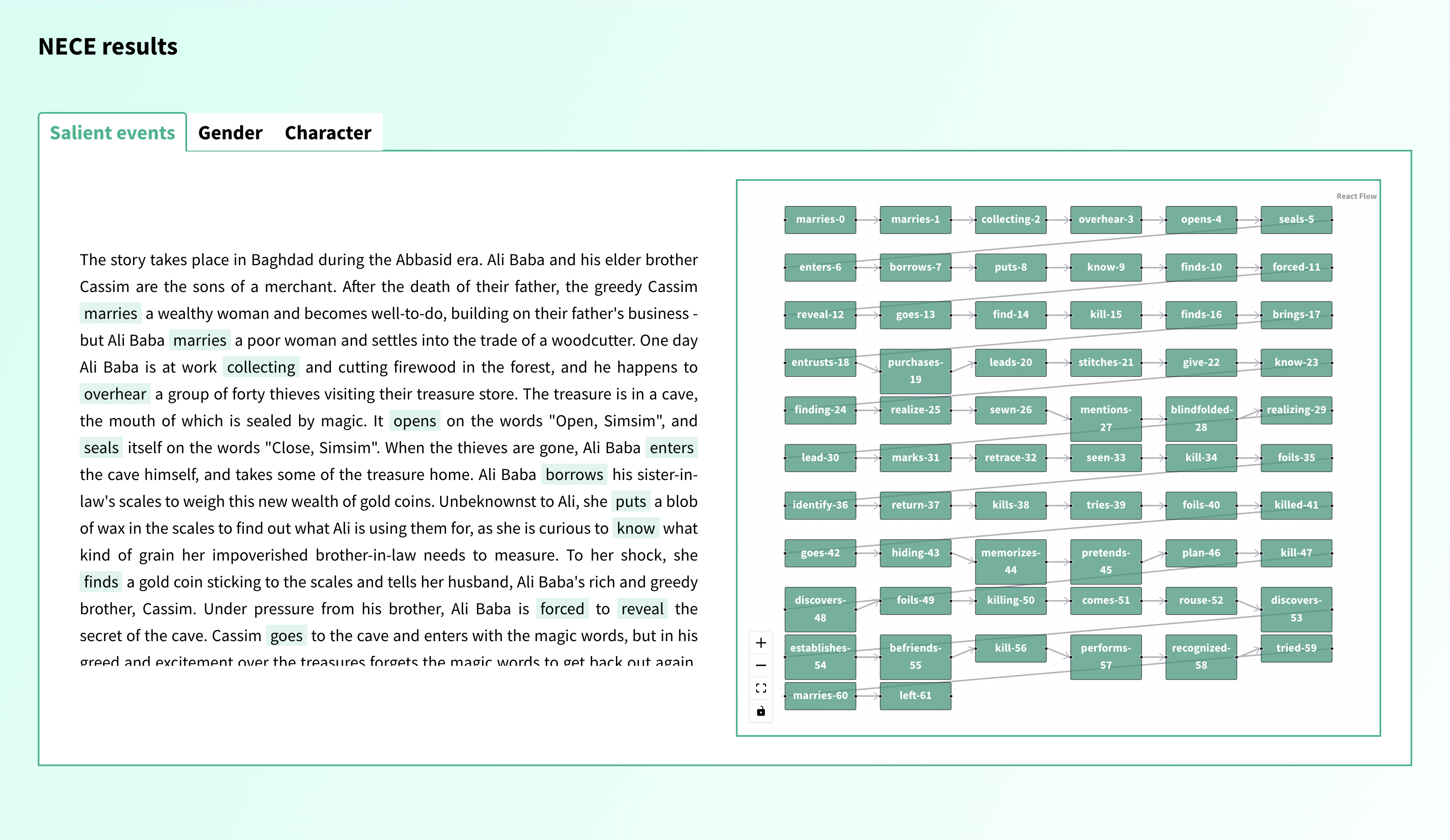}
    \caption{The chain of salient events highlight key plots associated with main characters of the narrative.}
    \vspace{-0.15in}
    \label{fig:demo_v1}
\end{figure}

\noindent \textbf{Temporal Ordering} models are also key component of \textsc{nece}; we tested SOTA TempRel model \textsc{econet} ~\cite{Han2021ECONETEC}, finetuned on BERT to anchor events onto temporal axis and resolve ordering by event start time(MATRES dataset~\cite{NingWuRo18}). Related TempRel models include TCR~\citep{ning-etal-2018-joint} which trains on annotated data plus ILP constraints, and ROCK~\citep{zhang2022causal} which finetunes RoBERTa~\citep{zhuang-etal-2021-robustly} on a 400k large automatically curated temporal dataset. DocTime~\citep{mathur-etal-2022-doctime} combines the training of a dependency parser and a TempRel predictor by formulating an edge prediction objective to generate temporal dependency graphs. Generative LLMs like ChatGPT and Flan-T5~\citep{Chung2022ScalingIL} may also be prompted to make temporal predictions. 

 % describes a novel salience aware algorithm to find events central for plot development, produce disconnected event chains for Natural Language Understanding tasks. 

% EventPLUS~\citep{Ma2021EventPlusAT} leverages TempRel models to create sentence-level temporal graphs; 

% trained on the  dataset. Existing works use temporal models, like ECONET or TEAR~\citep{han-etal-2019-joint}, to construct disconnected local event graphs or event chains created by flattening temporal relations graphs~\cite{Ma2021EventPlusAT,Dror2022ZeroShotOE,Zhang2021SalienceAwareEC}.  The intuition for their disconnected structure is that not events happened in the same temporal axis~\citep{NingWuRo18}, and are thus, comparable. We decided to adopt the compromised assumption that events are on the same temporal axis, and are comparable, due to the need to simply narrative representation, and that no document level temporal relation prediction model exists to produce coherent temporal order graphs. We propose an insertion-sort algorithm to leverage pair-wise temporal relations to construct an holistic event chain. 

\noindent \textbf{Online Interfaces} EventPlus~\citep{Ma2021EventPlusAT} features sentence-level online temporal reasoning interface. StoryAnalyzer~\citep{mitri} website offers narrative analysis capabilities built upon CoreNLP~\citep{manning-etal-2014-stanford}. \textsc{Nece} expands on both to offer narrative central investigation of temporal event flows, at document level. Our user-friendly website/interfaces can be used for either industry adoption or education purposes. 

% StoryAnalyzer in semantic feature capabilities, such as main character identification and event/participant highlighting. %etc. 
% \textsc{Nece} offers distinct support for salient events identification, document-level temporal events ordering, and %graphical
% narrative event chain visualization.

% BookNLP is a popular package for narrative processing, with strong performance in coreference resolution, character clustering, supersense tagging specifically trained for narrative documents. \textsc{Nece} uses BookNLP as the main information extraction engine. 
%~\citep{naseem-etal-2022-docamr} offers a document-level semantic parsing algorithm that summarizes the detailed document semantic information in a gigantic graph. 

% \noindent \textbf{Gender Fairness} in storybooks is an important motivation for the \textsc{Nece} toolkit. We created an independent interface for gender event chains and offer automatic gender bias analysis based on odds ratio as described in ~\newcite{sun2021men}. Representation of women in fairy-tales and gender power dynamics have been studied and examined by social researchers~\citep{lieberman1972someday,shaheen2019exploring}. Traditionally, this analysis often requires extensive manual coding in order to arrive at results. We hope \textsc{Nece} can reduce such burden through its event-based automatic bias analysis pipeline. 

\begin{figure*}[tbp]
    \centering
    \includegraphics[width=1.0 \linewidth]{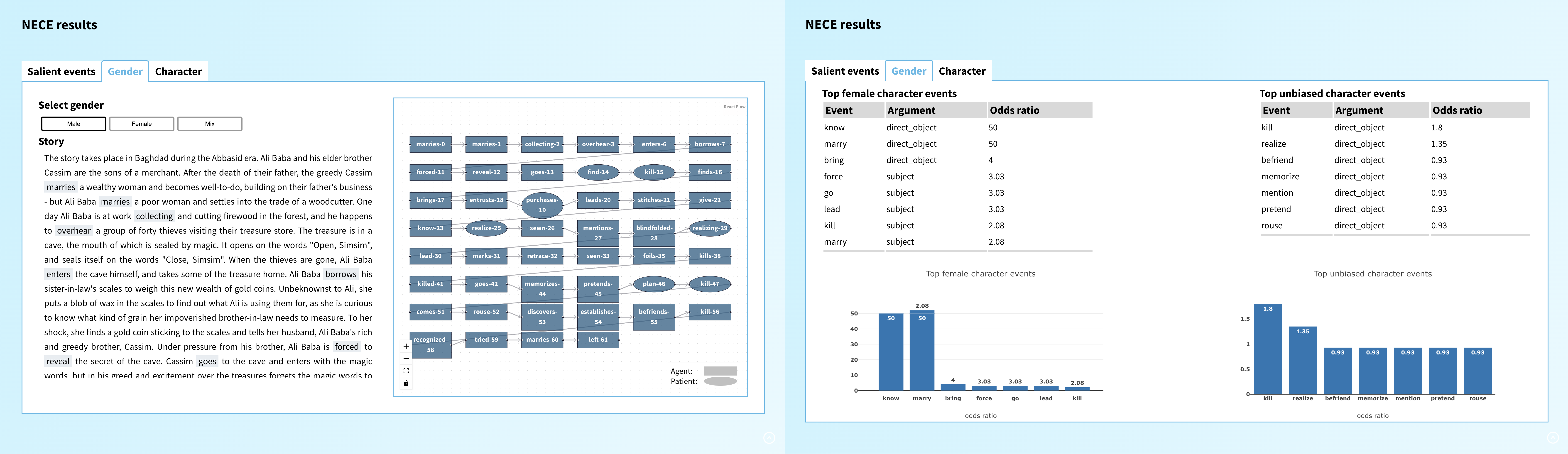}
    \caption{The gender interface displays event chains for Male, Female, and Mixed groups; scrolling downward button will display statistics and plot of bias analysis based on odds ratio.}
    \vspace{-3mm}
    \label{fig:demo_gender}
\end{figure*}

\section{Interactive Interface}
\label{sec: sys}

% \subsection{Online Interface}

\noindent \textbf{Salient Event Chain} Figure~\ref{fig:demo_v1} shows the first window of the \textsc{Nece} toolkit, offering the full salient event chain for the input narrative text. Filtering out non-salient event can reduce distracting items and makes the resulting chain more condensed and informative. This interface offers bird's-eye view of full event flow of the document, in the temporal order of their occurrence.  
% Our online interface is a salient event chain extraction and visualization tool, which is demonstrated . 
% In order to provide a more focused and comprehensive summary of the story, we define salient events as those that are relevant to the main plot and associated with the main characters of the narrative.To that end, we have selected only these salient events to be included in the event chain. The events are organized in temporal order, so that they are displayed in the order in which they occurred. The resulting salient event chain serves as a concise representation of the overall event chain of the entire story.

\noindent \textbf{Gender Event Chain} Figure~\ref{fig:demo_gender} shows the second interface of \textsc{Nece} that further organizes the salient event chain by gender association. Users can select and view event chains associated with either male or female characters or plural gender group of characters. This division is motivated by concerns for gender fairness in children's books~\citep{mccabe2011childrensbooks}. We build an automatic bias visualization based on odds ratio; we also used \textsc{nece} to perform a temporal anchored bias analysis to shed light on its application potential.

\noindent \textbf{Event Chain for Characters} is the third interface of \textsc{Nece} that displays salient event chains for the main characters (Figure \ref{fig:demo_character}). Character is another key characteristic of narrative text; event chain for each main character gives user easy access to plots involving their interested character. Main characters are decided based on combined name mention and pronoun mentions; participant role of characters is also differentiated by the shape of the event module: a square indicates an agent/subject participant, and an oval shape implies a patient/direct\_object participant of an event.

\begin{table}[t]
\centering
\resizebox{0.5\textwidth}{!}{%
\begin{tabular}{lcccc}
\toprule
Approach            & Dev-Prec.   & Dev-Recall & Test-Prec & Test-Recall \\ \midrule
No-Filtering          & 0.800     & 1.000        & 0.695         & 0.979 \\
Auxiliary-Verb-Filtering     & 0.910     & 1.000        & 0.842         & 0.979 \\
Salience-Filtering        & 0.921     & 0.978       & 0.857       & 0.979 \\ \bottomrule
\end{tabular}
}
\caption{Evaluation of no-filtering, auxiliary-verb-filtering, and salience-filtering approaches. The testset is a small set of 5 expert-annotated stories; all approach report near perfect recall indicates that our SRL-based event extraction approach is successful in identifying events. }
\label{tab:salience_filtering}
\end{table}

\begin{table*}[]
\centering
\small
\renewcommand{\tabcolsep}{1.8mm}
%\resizebox{\textwidth}{!}{%
\begin{tabular}{@{}llllllllll@{}}
\toprule
\multirow{2}{*}{\textbf{Method}} &
  \multicolumn{2}{c}{\textsc{Before}} &
  \multicolumn{2}{c}{\textsc{After}} &
  \multicolumn{2}{c}{ \textsc{F1}} &
    \multicolumn{1}{c}{ Kendall's $\tau$} &
  \multicolumn{2}{c}{\# \textsc{TrueLabel}}  \\ \cmidrule(l){2-9} 
 &
  \multicolumn{1}{c}{prec} &
  \multicolumn{1}{c}{rec} &
  \multicolumn{1}{c}{prec} &
  \multicolumn{1}{c}{rec} &
  \multicolumn{1}{c}{micro} &
  \multicolumn{1}{c}{macro} &
    coefficient &
  \multicolumn{1}{c}{before} &
  \multicolumn{1}{c}{after}  \\ \midrule
\textsc{random-order} (\textit{ref.})   & 0.729       & 0.653        & 0.203            & 0.260            & 0.557            & 0.456  & 0.709             & 262              & 85             \\
\textsc{text-order}    & 0.751              & \textbf{1.0}              &   0.0           & 0.0             & 0.750    & 0.428 & 0.725   & 262           & 85         \\
\textsc{flan-t5-large(fs)}    & 0.777             & 0.901              & 0.270             & 0.210             & 0.733            & 0.549       & 0.724        & 262               & 85              \\
\textsc{chatgpt}    & 0.763             & 0.737              & 0.274             & 0.318             & 0.631            & 0.519       & 0.717        & 262               & 85              \\
\cmidrule(r){1-1}    
\cmidrule(r){1-10}    
% (\textbf{ours}) \textsc{greedy}+\textsc{econet}      & \textbf{-}              &-              & \textbf{-}    & \textbf{-}    & \textbf{-}             & \textbf{-}     & 0.735         & 262               & 85                 \\ 
(\textbf{ours}) \textsc{greedy}+\textsc{econet}      & \textbf{0.933}              &0.858              & \textbf{0.689}    & \textbf{0.807}    & \textbf{0.848}             & \textbf{0.810}     & \textbf{0.735} | 0.728        & 262               & 85                 \\ 
\bottomrule
\end{tabular}%}
\caption{ \textsc{before, after, f1}, measures the quality of pairwise temporal relation extracted; one story consists of dozens of pairwise relations; Kendall's $\tau$ on the other hand, measures the holistic quality of temporal event chain, where one chain is created per story. \textsc{random-order} refers to random ordering of two events; \textsc{text-order} is the order that events appear in text. Best scores are \textbf{emboldened}. Due to imbalanced dataset labels favoring \textsc{before} relations, taking text-order gives good accuracy for this category. The authors have tried various prompting methods to use \textsc{chatgpt} for temporal classification, but with little avail. \textsc{flan-t5-large(fs)} adopts the same few-shot prompting as ChatGPT and gives comparable performance to the text-order heuristics. \textsc{greedy}+\textsc{econet} uses a greedy approach to construct event chain, only compare each event with its neighbors; on Kendall's $\tau$, we also report the result from \textsc{harmonypred}+\textsc{econet} of 0.728, which is lower than the Greedy approach overall. Though it performs better in some stories with more complex temporal timeline. }

\label{tab:main_results}
\end{table*}

% the same pairwise f-scores as the row below, but during event chain creation, it took a greedy approach to only compare with its adjacent events; it assumes a text-order, and swaps order only when an \textsc{after} relation is predicted. Overall, (\textsc{harmony}+\textsc{econet}) benefits from looking at larger context, and scores higher in Kendall's $\tau$.}

% Researchers and developers can use \textsc{Nece} through a Python package available on GitHub to output the full processing results of a narrative document. The general public can also easily access the toolkit through an online interface to process narrative texts of their choice and interactively find visualizations, analyses, and summaries of the uploaded document. 

\section{Event Chain Extraction Algorithms}
\label{sec: method}

%\subsection{Extract Narrative Salient Events}
\subsection{Salience Scoring}
\label{sec:salientEvent}

Salience identification is an algorithm that underpins the utility of event chain pipeline. Unfiltered event chain, taking all verb events from a semantic role labeling (SRL) model \citep{Gardner2017AllenNLP} will result in a flood of auxiliary verbs(is, are, am) and stative verbs(has, had) that not only distracts our temporal ordering model, but also make reading the event chain less interesting for users. 

% The process of creating an event chain begins with identifying all relevant events in a narrative. We use semantic role labeling (SRL) model \citep{Gardner2017AllenNLP} to extract verb events and their participants from the text. While we recognize that not all events are necessarily verbs, we adopt a narrow definition of events as verbs \citep{Han2021ECONETEC,NingWuRo18,zhang2022causal}.

Table ~\ref{tab:salience_filtering} shows that removing auxiliary verbs alone improves the salience detection by large margin. We further apply an additional layer of filtering to consider multiple dimensions including word frequency, character reference, and location reference, which are inspired by \citep{jindal-etal-2020-killed}. We trained an inverse document frequency (idf) dictionary on 2000 articles from wikipedia dataset~\citep{wikidump} to calculate tf-idf scores for each event. In addition, we aggregate factors, such as whether argument-0 or argument-1 of event trigger involves main characters or locations.  The threshold for salience grouping is relatively low to get high recall. The specific weights on components of salience score is obtained through grid-search on a 3-story dev set. Our final salience filtering algorithm scores a modest improvement on precision over auxiliary filtering, with no cost on recall as shown in Table\ref{tab:salience_filtering}.

 \subsection{Character Feature Extraction}
Organizing event chains by gender and character groups is dependent upon successful tagging of those features. We use a modified version of BookNLP \cite{Bamman2014ABM} to extract and cluster characters; each event is then tagged with characters as either agent participant or patient participant. By counting the total name and pronoun mentions for each character, we assign primary, secondary, and tertiary importance status to characters. 

Having events tagged with character associations, we can then create their gender tags by predicting the characters' gender identity. The predictions are derived from co-reference resolution to look up characters' associated pronouns. We also created a dictionary of common names for different genders, which is directly applicable when character names are explicitly gendered. Due to limitations of the current method, we only offer an imperfect gender categorization as female, male, group/non-binary, or unknown (See Limitation~\ref{sec:limits}) 

% In cases when the name of the character is explicitly gendered, we use direct mapping to assign a gender.

% looking up characters' associated pronouns using co-reference resolution and POS-tagging results. 

Having events tagged with gender roles, we extract odds ratio to illustrate potential gender bias in story, as in ~\newcite{sun2021men} and \newcite{ monroe_colaresi_quinn_2017}. For example, in a given story, the occurrence of the event ``kill'' has an odds ratio of four for male vs. female characters. This means that male characters are four times more likely than female characters to be involved in killing. We apply a common correction, Haldane-Anscombe~\citep{Lawson2004OddsRatio}, to account for cases in which one group has no observed counts of an event. 

% \noindent \textbf{Bias Analysis} We use the odds ratio as in~\citep{Sun2021MenAE, monroe_colaresi_quinn_2017} to illustrate potential bias in event for different gender groups. 

% Similarly, we tag gender roles for each event using our custom gender role prediction algorithm; which predicts a character's gender by lookup its associated pronouns through co-reference resolution; we also use direct mapping when the name of character is explicitly gendered. 

% the corresponding pronouns  through co-reference resolution and POS-tagging(NLTK). We also use direct mapping when the character name's gender implication is explicit. 

% extracts and clusters characters and performs co-reference resolution to link pronouns to characters. Primary, secondary, and tertiary characters are identified by counting total number of their name mentions and pronoun mentions. 

% \textsc{\sys} incorporates numerous narrative features to enable character-based event chain and gender event chains. 
 
% \noindent \textbf{Character Clustering and Coreference Resolution} 

% We use the BookNLP output to count the combined name mentions and pronoun mentions for characters to arrive at their relative importance category, and a high-level categorization as primary, secondary, and tertiary characters.

% \noindent \textbf{Gender Prediction}: 

\subsection{Document-level Temporal Event  Ordering}
\label{sec:eventordering}
\noindent \textbf{Pairwise TempRel.} Temporal orders are key features of narrative event chains. We first obtain pairwise temporal relations (TempRel) of events by using \textsc{econet} \cite{Han2021ECONETEC}. \textsc{econet} is one of  the \textsc{sota} neural temporal models using a BERT architecture ~\citep{Devlin2019BERTPO} and is trained with a joint-objective of temporal relation mask-filling and contrastive loss from discriminating corrupted temporal tokens.

% While the temporal ordering model ECONET can predict the pairwise relations between two events in the same sentence or adjacent sentences, it does not directly compute a ranking, and it doesn't support comparing events that are located far away from each other in text. 

\noindent \textbf{Greedy Event Chain Creation.} Pairwise relations between adjacent events do not directly compute a ranking of events. So, we propose an ILP(Integer Linear Programming) constraint which first assumes a textual order of events, and only change order of adjacent events based on pairwise relations from model prediction. The process can be also viewed as an insertion sort: let X be a sorted sequence of events with length $n \geq 0$ and let $x$ be a new event to be inserted into X, the algorithm consists of the following steps: Initialize a hashmap $T$ of temporal relations, where $T[i,j]$ represents the relation between events $i$ and $j$. Populate $T$ with pairwise relations between neighboring events in document: $T[i,i+1] = R$, where $R$ is the relation between events $i$ and $i+1$. The pairwise relation is predicted by the \textsc{econet}  model \citep{Han2021ECONETEC}. After having a hashmap of known relations, we use the insertion sort algorithm as in \textbf{Algorithm 1} to recursively insert a new event into a sorted chain to generate the final ordered event chain.

\begin{algorithm}
\label{alg:insertionSort}
\caption{Insertion Sort of Event Chain}\label{alg:cap}
\begin{algorithmic}
\Require Event sequence $X$ and a new element $x$, a hash-map $D$ containing pair-wise relations between events.
\Ensure $X$ is non-empty and sorted in temporal order, $x$ has text-order after all elements in $X$.
\State $b \gets X[-1]$
\While{$b$}
\If{$x$ comes after $b$ according to $D$}
\State Insert element $x$ right after $b$ in $X$
\State \textbf{return} $X$
\Else
\State $b \gets$ the element right before $b$ in $X$
\EndIf
\EndWhile
\State Insert element $x$ at the first index of $X$
\State \textbf{return} $X$
\end{algorithmic}
\end{algorithm}

% $T[i,j] = T[i,k] \circ T[k,j]$, where $\circ$ is a binary relation such as "before" or "after."

% We further enhance the above algorithm by expanding the scope of TempRel prediction: only considering pair-wise relations between immediate neighbors limits the model's capability in understanding longer term temporal dependencies. 
\noindent \textbf{Harmony Predict: Window-Sliding.} While greedy method above works well in scenarios where textual order heavily overlaps with temporal order, the greedy algorithm considers only an event's immediate neighbors during chain ordering; in order to capture longer dependencies, we \textbf{increase the window size to 5}: for each Event-i, pairwise TempRels are also computed for Event-${i-2}$ and Event-${i+2}$. We expand the ILP rules to include the transitive property of temporal relations to deduce additional relations in $T$ that are beyond the size-5 context window; for instance, if $T[i,k]=T[k,j]=``After"$, then $T[i,j]:=``After"$.  However, conflict may arise due to co-existence of transitive-inferred labels and model predicted labels. Harmony Predict apply the following ILP rules to reconcile differences: \textbf{1)} Text-order is preferred when the relationship between two events is unpredicted; \textbf{2)} transitive rules are applied to infer new relations from existing relations. \textbf{3)} in case of conflicts, transitive rules take precedence over the window-size 5 model's predictions, since \textsc{Econet} is more reliable for adjacent events.

\section{Evaluation}
\label{sec:fairytale}

% \subsection{Experiment Setup}
% \noindent \textbf{Dataset} Experiments are conducted on a subset of the FairytaleQA dataset, a corpus of 278 open-source fairy-tales~\citep{xu-etal-2022-fantastic}. For temporal ordering task, we asked professional in-house annotation team to rank extracted events by the temporal order that they occur, to create a validation dataset of pairwise temporal relations(tempRels) and ordered chains. Since event chain requires strict ordering, Vague tempRel and Simultaneus tempRel are no longer available. We used imperfect heuristics as listed in Appendix to resolve ambiguity. The testset for narrative features extraction are also sampled from FairytaleQA dataset. We asked annotators to identify salient events from generic events, and asked about the character association of events are correct, and also asked annotators to identify gender associated with events. 

\noindent %\textbf{Dataset} 
We carry out a series of experiments to understand the performance of \textsc{Nece}. An in-house annotation team annotated event saliency, gender and character association of stories taken from the FairytaleQA dataset~\citep{xu-etal-2022-fantastic} using instruction and interface shown at Figure~\ref{fig:annotation2}. For temporal ordering (tempRels) annotation, we sampled 4 stories from both FairytaleQA and 4 from cnn-dailymail~\citep{See2017GetTT} dataset to diversify the domain(Fairy tale stories mostly adhere to textual order; while news stories have more complex timelines). The detailed annotation guideline and the example annotation interfaces can be found in Appendix~\ref{sec:annotation}.

\subsection{Temporal Ordering}
\label{sec:eval_temporalOrder}
% \footnote{The ECONET model is trained on the MATRES dataset; see https://github.com/PlusLabNLP/ECONET.}
\noindent \textbf{Pairwise Temporal Ordering.} \textsc{Nece} employs \textsc{econet}~\citep{Han2021ECONETEC} for TempRel prediction between pairwise events; \textsc{econet} is finetuned on small-size (36 documents) but expert-annotated MATRES dataset~\citep{NingWuRo18}.  We added random prediction as a reference and text-order of events as a strong heuristic-based baseline. We also included few-shot prompted flan-t5-large~\citep{Chung2022ScalingIL} and ChatGPT(gpt-3.5-turbo) to explore the temporal reasoning capability of instruction-finetuned LLMs. Pairwise temporal relation are measured by F1 scores, as shown in Table \ref{tab:main_results}.

\noindent \textbf{\textsc{llm}.} Generative AI spearheaded by \textsc{chatgpt} has made ripples in many domain by demonstrating strong zero-shot or few-shot capabilities. We tried to leverage gpt-3.5-turbo using few-shot examples and dialog style prompts to extract temporal predictions from it, though with little success. The Table~\ref{tab:chatgpt_comparison} in Appendix gives detailed example of its failure in generating meaningful temporal prediction. It tends to generate one label(yes or no), or always the first event if prompted differently to choose the earlier event. Its poor performance is reflected in Table~\ref{tab:main_results} as low f-scores; we then used the same prompts and set-up as \textsc{chatgpt} on Flan-T5-large~\citep{Chung2022ScalingIL}, which gives promising results close to text-order.

\noindent \textbf{Document-level Temporal Event Chain.} To benchmark the quality of constructed event chain, we use the Kendall's $\tau$ coefficient, which measures the similarity of the orderings of the data~\citep{Kumar2010GeneralizedDB}. A higher Kendall's $\tau$ coefficient indicates better alignment between predicted chain and the real chain of event. In a holistic view, our \textsc{greedy+econet} approach not only scores high in pairwise prediction, but also has the closest event chain to gold. \textsc{harmony+econet} which uses the window-sliding method to construct event chain have a lower average score in Kendall's $\tau$, but scores higher in some stories with complex temporal timelineTable ~\ref{tab:main_results}. 

% is defined as the number of swaps required to convert the second sequence into the first using bubble sort

\noindent \textbf{Discussion}  In Table \ref{tab:main_results}, we find the \textsc{Text-order} model a strong baseline in terms of micro-F1 due to imbalanced labels favoring \textsc{before} relation. The result of \textsc{flan-t5}'s few-shot performance is surprisingly good, achieving 0.777 and 0.9 in precision and recall for \textsc{before} label, especially compared to \textsc{chatgpt}. Temporal ordering task typically involves complex definition to resolve overlap in time-span and multi-axis scenarios, making direct knowledge transfer difficult.  Our \textsc{Econet} based methods outperform the strong baselines in most of the metrics, including pairwise f-score and chain-level Kendall's $\tau$ measure;

\begin{table}[t]
\centering
\resizebox{0.5\textwidth}{!}{%
\begin{tabular}{llll}
\toprule
Event-chain            & accuracy   & macro-f1 & \# samples \\ \midrule
Character Resolution     & 0.872     & -        & 188         \\
Gender Resolution        & 0.974     & 0.951       & 188       \\ \bottomrule
% Temporal-Order               & 0.94.3            & 196         \\ 
\end{tabular}
}
\caption{Evaluation of character resolution, and gender resolution algorithms; we report the accuracy and macro-f1 for gender resolution; however, only accuracy is reported for character resolution because number of character classes is not fixed across different stories.}
\label{tab:event_chain_eval}
\end{table}
\subsection{Feature Extraction}
\label{event_chain_table}
We evaluate three key narrative features that are crucial to construct event chains. Salience feature is useful to filter out redundant and generic events.  Gender and character tagging allows us to organize event chains by their gender and character association.  Result for salience filtering is shown in Table~\ref{tab:salience_filtering} and discussed in section~\ref{sec:salientEvent}. Gender prediction, which is a relatively easy task solvable through gendered pronoun identification and co-reference resolution, enjoys high accuracy as shown in Table~\ref{tab:event_chain_eval}.  Character identification is a slightly more challenging task due to the large number of possible characters in a narrative;   its performance also depends on successful character clustering. We still achieve good performance thanks to the robustness of the BookNLP toolkit~\citep{Bamman2014ABM} in narrative domains. 

% Salience detection and gender tagging is measured by both accuracy and macro average F1 score. The character association task is only measured with accuracy as different stories contain different numbers of characters.

% % To evaluate the performance of our algorithm for salient event identification, we conduct experiments on a corpus of stories. 
% % The stories contain hundreds of events, making manual coding on the four dimensions a laborious task. Therefore, we adopt a precision-based method to crowdsource judgement for model prediction.

% \noindent \textbf{Result Analysis} 

% \begin{figure}[tbp]
%     \centering
%     \includegraphics[width=0.4\textwidth]{images/6_2_bigrams_possession.png}
%     \caption{When the odds ratio is high for a particular gender, it suggests that this gender has a greater probability of experiencing the event}
%     \vspace{-0.15in}
%     \label{fig:bias_analysis}
% \end{figure}
% \begin{figure}[h]
%     \includesvg[width=1.0\linewidth]{images/6_0_analysis_units.svg}
%     \caption{Proportion of Significant Gender Bias by Analysis Unit.  When looking at the location of an event within a character's narrative arc, female characters have more biased events in the beginning of their arcs while the bias for male characters is fairly consistent throughout all three sections of their arcs. }
%     \label{fig:analysis_units}
% \end{figure}

\begin{figure}[tbp]
    \centering
    \includegraphics[width=0.48\textwidth]{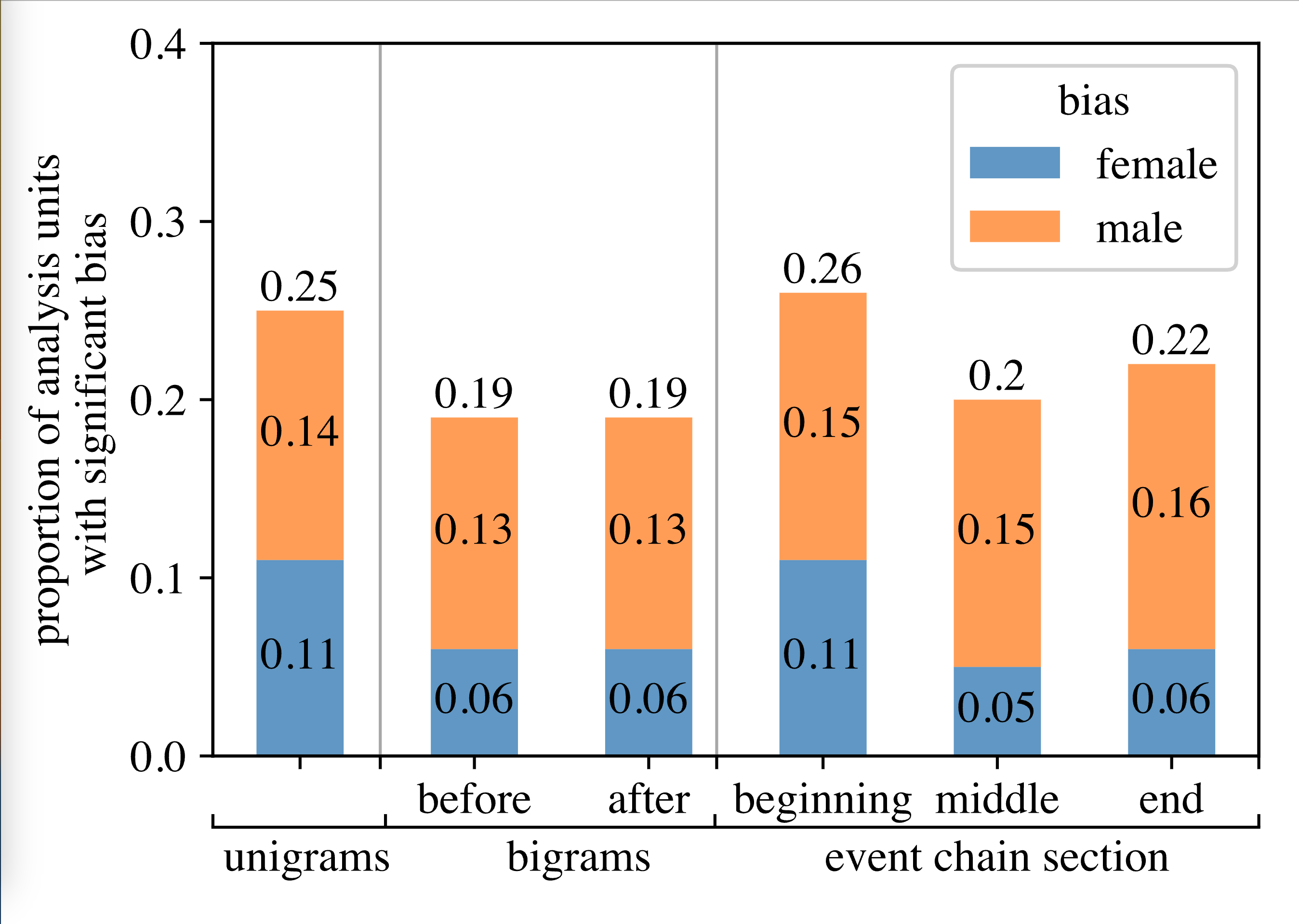}
    \caption{Proportion of Significant Gender Bias by Analysis Unit.  When looking at the location of an event within a character's narrative arc, female characters have more biased events in the beginning of their arcs while the bias for male characters is fairly consistent throughout all three sections of their arcs.}
    \vspace{-0.15in}
    \label{fig:demo_character}
\end{figure}
\subsection{Application in Temporal Event Chain-Anchored Gender Bias Analysis}
\label{sec:biasalaylsis}
\textsc{Nece} can be used to investigate gender bias that arise from the order in which events happen. We tested \textbf{Bigram Event Comparisons} to locate such biases within the FairytaleQA dataset~\citep{xu-etal-2022-fantastic}. We integrated temporal ordering as a dimension of bias, and we compared unigram, bigram bias, at different temporal locations of the narrative(beginning, middle, end); we also investigated the possibility of bigram bias, as the prior event(before) or the later event(after).

The problem is formulated as following: Bigrams (chains of two events) are extracted from each event chain. For example, a bigram could be (``cry'', ``have''); they are further grouped into custom supersense categories (e.g., being mapped to the typical verbNET supersense categories such as  ``emotion'' or ``possession''). We then selected one supersense category, such as \textit{possession} as the anchor, and calculated the odds ratio for all events that occur prior to the anchor. An example analysis is shown in Figure \ref{fig:analysis_units}, biases at before and after locations seem to be even; however, there are significantly more biaes against female character at the beginning of the story than at the end. Anlaysis is based on 278 fairytales~\citep{xu-etal-2022-fantastic}.

\section{Conclusion}

% \sys can be a platform where other structured representation can be used to aid narrative comprehension. Think of some use cases;

\textsc{Nece} is a document-level narrative event chain extraction toolkit, built with neural temporal ordering models and novel character feature tagging algorithms. It boasts a user-friendly online interface along with built-in gender bias analysis capability.  \textsc{Nece} is built to serve event-based analysis of a wide range of narrative documents, including novels, short stories, fairy-tales, etc. Human evaluation is conducted on key algorithmic components of \textsc{Nece}, demonstrating its robustness and accuracy. While our system has been primarily designed for event chain extraction, its functionalities can be adopted for various downstream applications, including temporal anchored Bias Analysis. Classifer-based temporal ordering has its limits, though directly leveraging generative model for temporal classification is also non-trivial. Future work is needed to investigate generative AI's temporal ordering ability. 

% a novel system for extracting temporal event chains and character features from narrative documents.

% In conclusion, we present \textsc{Nece}, the first system online interface that extracts temporal event chains for narrative documents with built in functionalities for character feature extraction and analysis. Our pipeline is hosted on a public server which is powered by GPU to run fast inference for large neural models. We hope that \textsc{Nece} can assist research and work in event-based analysis or understanding of novels, short stories, fairy-tales, and other wide ranges of narrative documents. While bias analysis and Question-Answering are two example downstream tasks, the \textsc{Nece} system can be leveraged for many more potential applications.   

% can encourage more people to understand the bias under story construction, and explore the potential bias embedded in the temporal making of the narrative.  

\section*{Limitations}
\label{sec:limits}

The overall performance of \textsc{Nece} is constrained by performance of its component algorithms; temporal relation prediction models are trained on sentences level rather than document level; the pairwise predictions within a document span are also not always harmonious, and need ILP rules to resolve conflicts. Even we obtained all the correct pairwise relations, it could still be hard to build various temporal axis in the narrative, and assemble them into the final event chain. We explored using \textsc{chatgpt} and \textsc{flan-t5} models to make temporal order predictions, albeit with limited success. It merits future works to unlock the temporal capabilities of LLMs for temporal ordering task.

\section*{Ethics Statement}
\label{sec:ethics}

% Scientific work published at EMNLP 2022 must comply with the \href{https://www.aclweb.org/portal/content/acl-code-ethics}{ACL Ethics Policy}. We encourage all authors to include an explicit ethics statement on the broader impact of the work, or other ethical considerations after the conclusion but before the references. The ethics statement will not count toward the page limit (8 pages for long, 4 pages for short papers).

% The goal of this toolkit is to give authors, teachers, readers, and researchers the ability to surface potential gender bias in story texts. We hope that the results will extend and deepen the analysis and discussion of existing works as well as more thoughtful consideration to the construction of works in progress. 

In addition to being an event chain extraction toolkit, we also build in a gender bias analysis component. We make the normative assumption that any substantial, measured numerical difference between two groups is indicative of bias within a story. However, there is the caveat that any distributional imbalance found in single stories are normally not statistically significant. The odds ratio computed are only indicative of potential bias, not as conclusion of bias. Moreover, rather than conducting bias analysis on events level, it is better to first abstract events into verbNET supersense categories and conduct odds ratio analysis on supersense level, as done in Section~\ref{sec:biasalaylsis}. Moreover, we do not make any claims as to the polarity of bias nor does it contextualize such bias in the rich body of work of literary criticism, media studies, gender studies, or feminist literature. We are aware that numerical measures of bias can be used to obfuscate nuance or wave away concerns of harmful representation. We do not intend for our tool to replace qualitative analyses of stories, but rather supplement existing bias analysis frameworks. Lastly, we host a website to demonstrate our toolkit; however, due to the expense of host gpu-backed server, we only uploaded samples of processed contents to the website, but do not support online inference. We do not collect any user data from the website. 

% \section*{Acknowledgements}
% We'd like to thank Daniel Firebanks-Quevedo for his work on the gender feature extraction. We want to thank Nathan Good for providing advice and help in Website design and server scalability.

\bibliography{anthology,custom}
\bibliographystyle{acl_natbib}

\appendix
\label{sec:appendix}

\section{Annotations}
\label{sec:annotation}
Narrative feature annotations are conducted by in-house annotation team; the temporal ordering annotation is performed by both in-house annotator and cross-checked by authors of the paper to ensure adherence to the annotator rules. Our in-house annotators are compensated fairly in accordance with compensation law in the state of New York. 

\subsection{Temporal Event Chain Annotations}

\textbf{Task Formulation: } Given a passage of the story, a list of verb events, that can be mapped back to passage; the annotators are tasked to create a ranking of listed verb events by anchoring the events into the story timeline. The result will be a chain of event-ids, eg. 2->1->3->4->6->5 …. We ask annotators to put -the results into prepared excel sheets.

\textbf{Annotation rules}:
We trained the annotators on below listed rules to resolve ambiguities that arise during ambiguity. As pointed out by ~\citep{NingWuRo18}, not all events fall on the main story axis, with diverging hypothetical, negation, conversatoin timelines, or even past timelines that weren't able to be clearly located. This it makes the creation of one uniform event chain difficult, since there's a lot of ambiguities. During annotations, we apply the following rules to ensure a uniform way of resolving ambiguities;

%From , we use heuristics to calculate the four dimensions of engagingness scores. 
\begin{itemize}
\item \textbf{Hypotheticals:} Hypothetical events could be in the past or in the future; place the hypothetical timeline before its immediate neighbors if it were in the past; place the timeline after the immediate neighbors if it were in the future. 

\item \textbf{Contexts:} Some events are used to provide context for later conversations or plots, eg. a bridge \textbf{stood} above the river; make sure the contexts were temporally earlier than the events or conversation dependent on them; 

\item \textbf{Cause and Effect:} When there's cause and effect relations, such as, because, so, since, we should always put cause first, and then effects. 

\item \textbf{Quotations:} You can often infer the relations between events inside quotation and outside quotation; if it is ambiguous, annotate in text-order. 
\end{itemize}

\subsection{Narrative Features Annotations}
\begin{figure*}[!h]
    \centering
    \includegraphics[width=0.98\textwidth]{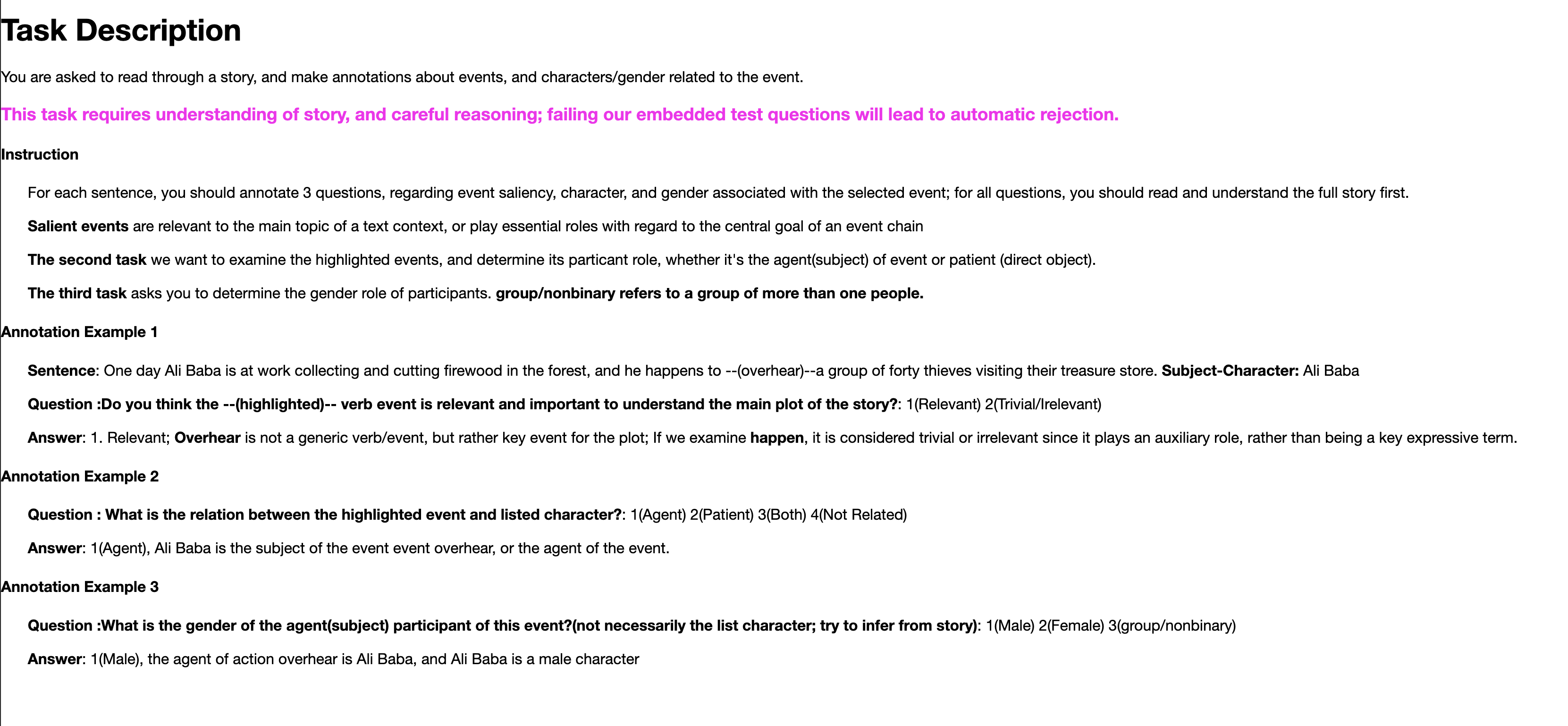}
    \caption{The screenshot of the task description for the annotation of extracted narrative features, which is used for annotation.}
    \label{fig:annotation1}
\end{figure*}

\begin{figure*}[!h]
    \centering
    \includegraphics[width=0.98\textwidth]{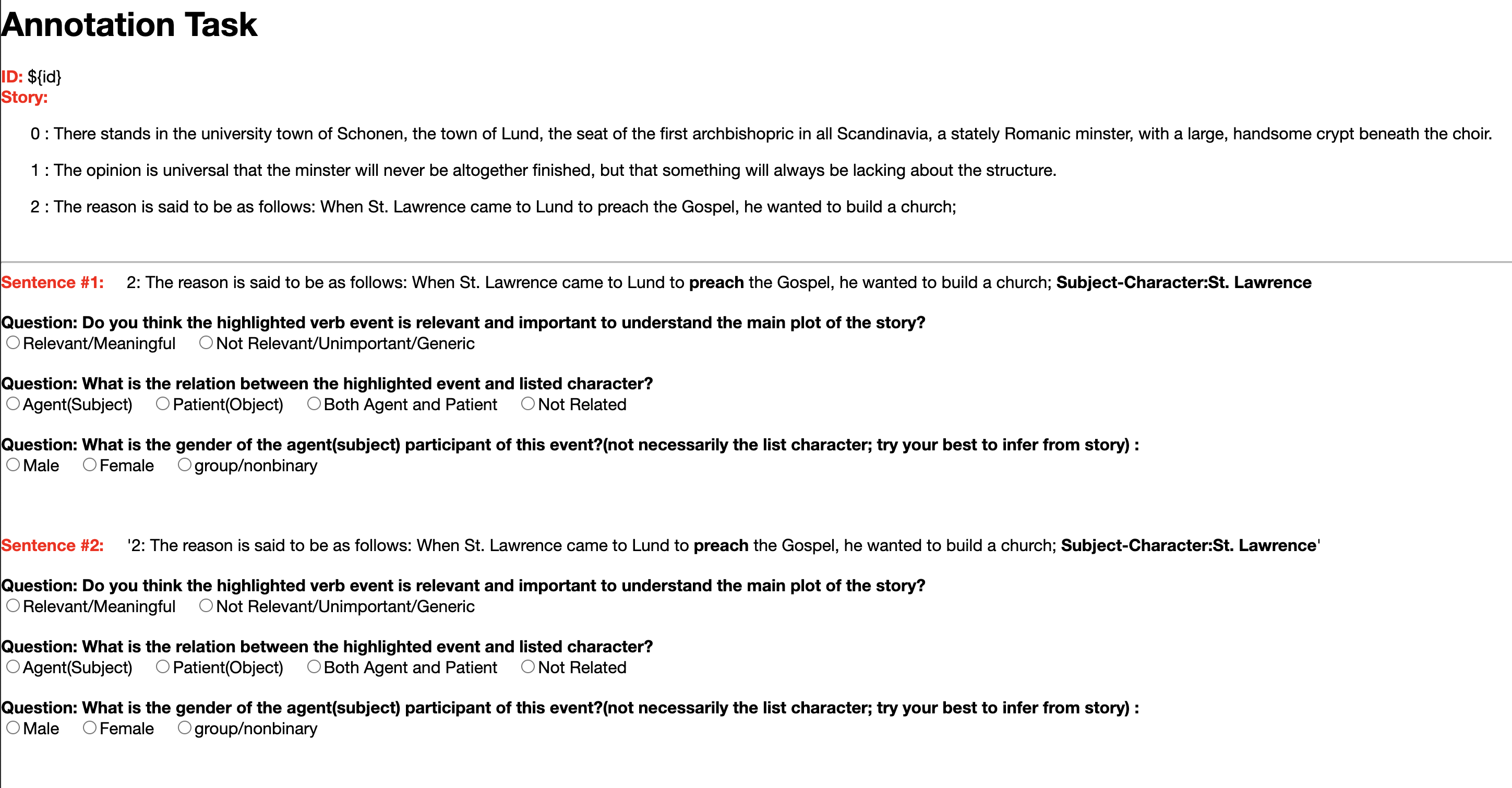}
    \caption{The screenshot of the annotation interface for an example short story.}
    \label{fig:annotation2}
\end{figure*}

Narrative features, including salience detection, gender role resolution, character role resolution are annotated using the following interface in Figure~\ref{fig:annotation1}~\ref{fig:annotation2}. We provide the entire story, and for each question, we provide the corresponding sentence and 3 questions about salience, gender, and character role.

\section{Use ChatGPT for temporal classification}

\begin{table*}[htbp]
    \centering
    \small
    \renewcommand{\tabcolsep}{1.8mm}
    \begin{tabular}{@{} p{5cm} p{5cm} c c c c @{}}
        \toprule
        \textbf{query1} & \textbf{query2} & \textbf{answer1} & \textbf{answer2} & \textbf{model\_pred} & \textbf{gold\_label} \\ 
        \midrule
        "in the following text: The call , which happened as President Barack Obama wrapped up his first presidential visit to Israel , was an unexpected outcome from a Mideast trip that seemed to yield few concrete steps.
        is it possible that event 'happened' happened before event 'wrapped' starts? Please only answer yes or no." &
        "in the following text: The call , which happened as President Barack Obama wrapped up his first presidential visit to Israel , was an unexpected outcome from a Mideast trip that seemed to yield few concrete steps.
        is it possible that event 'wrapped' happened before event 'happened' starts? Please only answer yes or no." &
        No & No & SIMUL & BEFORE \\
        
        "in the following text: The call , which happened as President Barack Obama wrapped up his first presidential visit to Israel , was an unexpected outcome from a Mideast trip that seemed to yield few concrete steps.
        is it possible that event 'happened' happened before event 'seemed' starts? Please only answer yes or no." &
        "in the following text: The call , which happened as President Barack Obama wrapped up his first presidential visit to Israel , was an unexpected outcome from a Mideast trip that seemed to yield few concrete steps.
        is it possible that event 'seemed' happened before event 'happened' starts? Please only answer yes or no." &
        No & No & SIMUL & BEFORE \\

        "in the following text: The call , which happened as President Barack Obama wrapped up his first presidential visit to Israel , was an unexpected outcome from a Mideast trip that seemed to yield few concrete steps.
        is it possible that event 'happened' happened before event 'yield' starts? Please only answer yes or no." &
        "in the following text: The call , which happened as President Barack Obama wrapped up his first presidential visit to Israel , was an unexpected outcome from a Mideast trip that seemed to yield few concrete steps.
        is it possible that event 'yield' happened before event 'happened' starts? Please only answer yes or no." &
        No & No & SIMUL & BEFORE \\
        
        "in the following text: The call , which happened as President Barack Obama wrapped up his first presidential visit to Israel , was an unexpected outcome from a Mideast trip that seemed to yield few concrete steps.
        is it possible that event 'wrapped' happened before event 'seemed' starts? Please only answer yes or no." &
        "in the following text: The call , which happened as President Barack Obama wrapped up his first presidential visit to Israel , was an unexpected outcome from a Mideast trip that seemed to yield few concrete steps.
        is it possible that event 'seemed' happened before event 'wrapped' starts? Please only answer yes or no." &
        Yes & No & BEFORE & VAGUE \\
        \bottomrule
    \end{tabular}
    \caption{Examples of ChatGPT model prediction and the prompts fed; Few-shot examples used is: "in the following text: Fidel Castro invited John Paul to come for a reason .is it possible that event 'invited' happened before event event 'come' starts? Please only answer yes or no. Yes'; in the following text: She says this puts the very existence of women 's families at risk .is it possible that event 'says' happened before event event 'puts' starts? Please only answer yes or no. Yes" }
    \label{tab:chatgpt_comparison}
\end{table*}
Table~\ref{tab:chatgpt_comparison} shows the query and resulting answer from ChatGPT, which totally fails to make reasonable reasoning of the temporal relationship between querid events. The few-shot prompt given is shown in the description of the table. We used the same few-shot prompts for ChatGPT as for Flan-T5 models, which give much better performance. ChatGPT, though powerful in the dialog and creative capabilities, is unfortunately not as capable in the temporal capability. We have tested other prompts as well, but still was not able to have ChatGPT give better temporal predictions. Our results should not be taken as conclusive, but it adds to the data points for us to interpret its temporal ability.

\section{Temporal Chain Cases Analysis}
\begin{figure*}[!h]
    \centering
    \includegraphics[width=0.98\textwidth]{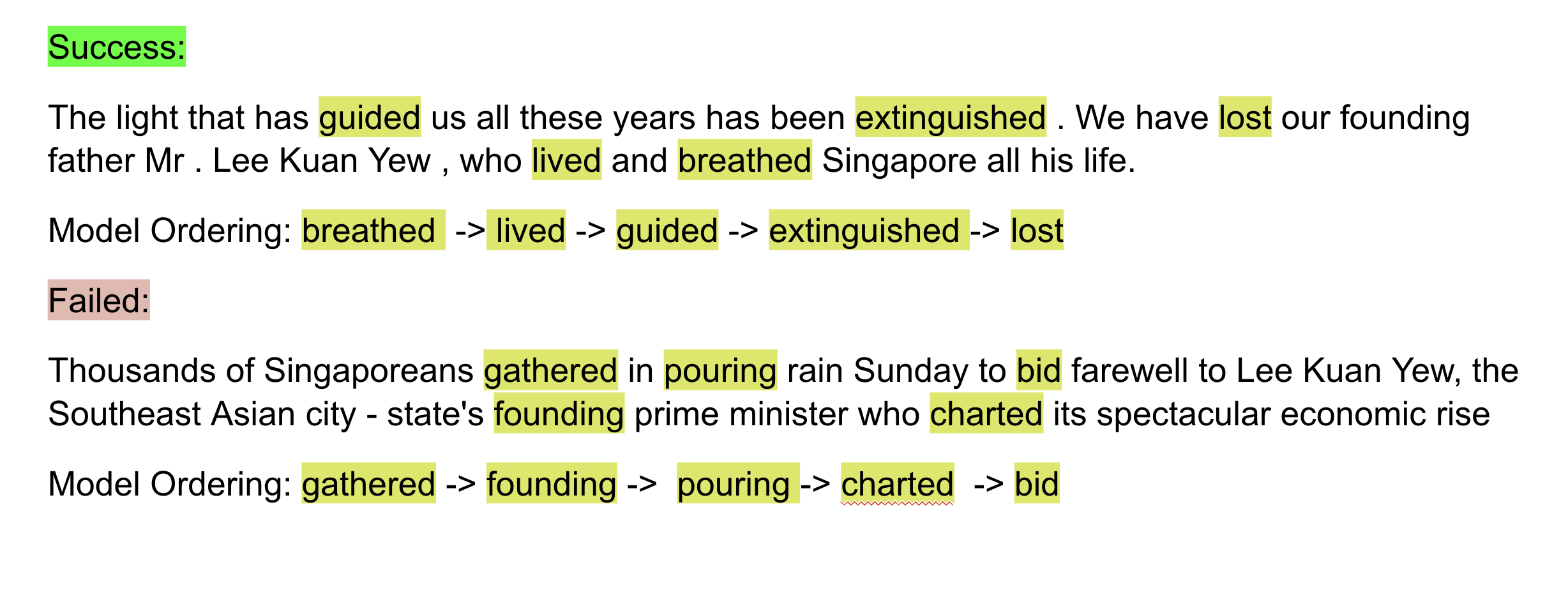}
    \caption{Example of successful model prediction by \textsc{nece}, and failure case, using a story from cnn-dailymail dataset; }
    \label{fig:temporal_example}
\end{figure*}

Figure ~\ref{fig:temporal_example} shows an example of successful ordering by \textsc{nece} and an example of failed ordering by \textsc{nece}. The current temporal reasoning models are really only trained on pairwise relations, without a systematic way to reconcile conflict and ensure a reasonable ordering in paraphragh or document. \textsc{nece} applies some ILP constraints to merge pairwise relations to chains, but is not the most robust solution. We hope future work can further explore the problem and propose better algorithms to holistically process the temporal ordering of narrative events. 

\end{document}